# Reinforcement Learning for Sociohydrology

Tirthankar Roy[1,*], Shivendra Srivastava[1], Beichen Zhang[2]

[1] *Department of Civil and Environmental Engineering, University of Nebraska-Lincoln*
[2] *School of Natural Resources, University of Nebraska-Lincoln*

*Corresponding Author: Tirthankar Roy (roy@unl.edu)

## Abstract

In this study, we discuss how reinforcement learning (RL) provides an effective and efficient framework for solving sociohydrology problems. The efficacy of RL for these types of problems is evident because of its ability to update policies in an iterative manner—something that is also foundational to sociohydrology, where we are interested in representing the co-evolution of human-water interactions. We present a simple case study to demonstrate the implementation of RL in a problem of runoff reduction through management decisions related to changes in land-use land-cover (LULC). We then discuss the benefits of RL for these types of problems and share our perspectives on the future research directions in this area.

## 1. Background

Sociohydrology has emerged as a new science of water (Di Baldassarre et al., 2019; Pande & Sivapalan, 2017; Sivapalan et al., 2014). Unlike traditional water resources management, where extrinsic scenarios are fed to a model to study the model response, in sociohydrology, those scenarios co-evolve intrinsically within the integrated human-water system. Thus, this explicit accounting for the two-way couplings between the human-water systems makes sociohydrology stand out as a novel approach to understanding the dynamics of human-water systems. Even though it has been widely acknowledged that sociohydrologic approaches can outweigh traditional approaches in terms of the realism they bring, only a handful of models have been proposed so far on this topic. Furthermore, these models are often conceptual in nature, primarily built for a specific problem at hand based on bottom-up approaches, which leads to a lack of generalization power.

Here, we argue that reinforcement learning (RL) provides a generalizable framework to efficiently and effectively model sociohydrology problems, which also functions across scales. In RL, the algorithms learn on the go by interacting with the environment based on the actions made and their subsequent rewards. This specific way of learning closely mimics the way adaptive water management decisions can be made, i.e., by constantly gathering feedback on the impacts of past interventions (decisions in this case) and updating the current interventions accordingly. We present a simple test case to establish the feasibility of this approach. We highlight the unique attributes of the framework, specifically focusing on the aspects of generalizability, transferability, scalability, data utilization efficiency, nonstationarity, and emergent phenomena. We also discuss which areas future research efforts can be directed to in order to bolster the use of RL in sociohydrology. Finally, we call for a community effort to make RL more mainstream to solve sociohydrology problems.

## 2. Test Case Demonstrating the Efficacy of an RL-based Sociohydrology Framework

The test case focused on land-use land-cover (LULC)-based management decisions meant to reduce runoff. We selected one of the Natural Resources Districts (NRDs) of Nebraska for our test case implementation. We precisely clipped the area near the northwestern side of Omaha, where we see growing development, making it ideal to test the hypothesis (RL can optimize land-use decisions). The land cover data was acquired from the National Land Cover Database (NLCD). Further, the National Oceanic and Atmospheric Administration (NOAA) precipitation dataset was used to calculate runoff. We considered seven different LULC categories (i.e., water, barren land, urban area, agricultural area, grassland, forest, and wetland) to design five distinct management scenarios, which include varied levels of these categories (**Table 1**). These scenarios were then used in the rational method—a simple runoff estimation equation that can consider the effects of LULC to calculate the runoff, as shown in **Figure 1**. In the next step, we applied the RL algorithm to enhance the current land-use pattern, followed by the calculation of runoff based on the optimized scenario. The results clearly demonstrate the benefits of scenario optimization with RL, as the resulting runoff from the RL-based optimized policy led to the minimum runoff when compared to runoff values produced by all different scenarios considered (**Figure 1**).

For the specific implementation, we applied a classic actor-critic RL framework, Proximal Policy Optimization (PPO), to learn the complex and sequential interactions between human activities (reflected through land use) and flood impacts (reflected through runoff changes). The main structure of PPO contains two neural networks: a policy function named Actor and a value function named Critic. The Actor controls an agent's actions, simulating the planning and decision-making processes in LULC management regarding flood risks. The Critic measures the quality of the agent's actions, and its outcomes will be input to update the policy. PPO works by encouraging actions that lead to higher rewards. Compared to the original Advantage Actor-Critic (A2C) algorithm, PPO improves the stability of training the Actor to learn the policy by constraining the updates of the policy network, which helps ensure a smoother and more reliable learning process (Schulman et al., 2017).

In our RL framework, action and state spaces were designed to represent human activities through changing the LULC and their impacts on runoff. The action here referred to the specific land use changes that the agent could undertake, such as altering agricultural practices, implementing nature-based solutions, or expanding urban development plans. The actions were critiqued based on their effectiveness in achieving the desired outcome of runoff reduction. The state represented the current configuration of the land use classifications affecting runoff. The PPO Actor network selects actions representing changes to LULC types and evaluates their impact on runoff using the empirical coefficients in the rational method—a simple runoff estimation equation that considers the effects of LULC. We carefully assigned the coefficients for the rational method, informed by our understanding of various land use and land cover classifications and their respective runoff generation responses as outlined in the literature (Shi et al., 2007; Sriwongsitanon and Taesombat, 2011). If the projected reduction met or exceeded the target, the action was deemed favorable; otherwise, it was considered inadequate. Through iterative learning, the Actor network was optimized through the gradient descent steps and updated its policy to favor actions leading to significant runoff reduction based on the Q-value predicted by the Critic network, facilitating the optimization of LULC strategies. The reward function was built based on the reduced runoff by the LULC change. Larger deficits in runoff values before and after the human modifications resulted in higher rewards, incentivizing the Actor network to learn effective policies to mitigate runoff. For this initial exploration of an RL application on sociohydrology, the research question was simplified to examine if the RL algorithm can learn how to reduce regional runoff through changing LULC types at the local level. As an evaluation, the outcomes from the PPO model after training in the LULC environment were compared to the five different management scenarios.

**Table 1:** Five different LULC scenarios designed to simulate runoff.

| LULC Types | Scenario 1 | Scenario 2 | Scenario 3 | Scenario 4 | Scenario 5 |
|---|---|---|---|---|---|
| Water | No change | No change | No change | No change | No change |
| Barren | 50% reduction | No change | 50% increase | 50% reduction | 50% reduction |
| Urban | No change | No change | No change | No change | No change |
| Agriculture | 10% reduction | 10% increase | 15% increase | 20% increase | 20% reduction |
| Grassland | 50% increase | 50% reduction | 20% reduction | 87.5% reduction | 75% increase |
| Forest | No change | 10% reduction | No change | 50% reduction | 100% increase |
| Wetland | 10% increase | No change | No change | 50% reduction | 100% increase |

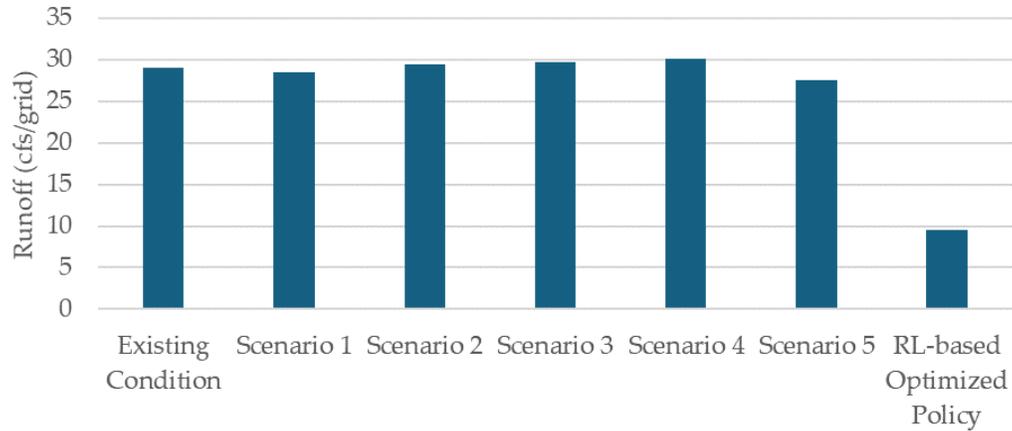

**Figure 1:** Comparison of runoffs from the existing conditions, designed scenario, and optimized (using RL) scenario.

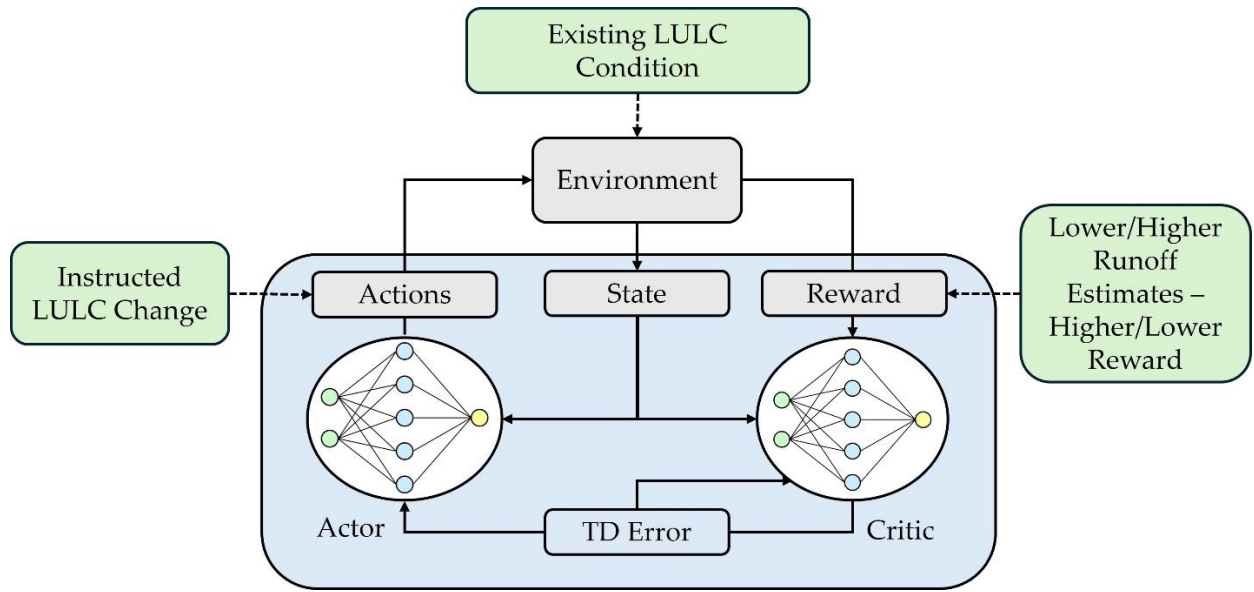

**Figure 2:** The core components of the RL-based sociohydrology framework.

The RL algorithm, in this case, updated the pixels such that most of the barren land, forest, grassland, and agricultural land have been converted to wetland in an effort to reduce runoff (**Figure 3**). This makes sense since wetlands are known to reduce runoff downstream (Srivastava et al., 2023). There have also been some minor changes. For example, some of the forest lands have been converted to agricultural land, and in one case, this change reversed, which is likely attributable to the spatial variability in these management interventions. Some of the grasslands also turned into forests. This clearly shows the potential of a framework like this in real-life water management. Note that the simulation can be made more realistic by adding constraints, such as some pixels not being updated, since that would be the case in real-world applications, where we often cannot just remove forests and construct wetlands. The imposed constraints can represent this reality in the scenario optimization process. Another potential avenue for improvement would be to incorporate spatial variation into the RL environment and the reward function, encouraging the Actor network to learn and produce policies with a lower cost of altering LULC types and capture a more realistic spatial representation of the adopted action, i.e., capturing a reasonable spatial variability in the action instead of a group of blanket actions.

|   | | | LULC after Action | | | | | |
|---|---|---|---|---|---|---|---|---|
| Land Cover | Total Pixel | Water | Urban | Barren | Forest | Grassland | Agriculture | Wetland |
| Water | 5 | 5 | 0 | 0 | 0 | 0 | 0 | 0 |
| Urban | 93 | 0 | 93 | 0 | 0 | 0 | 0 | 0 |
| Barren | 4 | 0 | 0 | 0 | 0 | 0 | 0 | 4 |
| Forest | 30 | 0 | 0 | 0 | 0 | 0 | 2 | 28 |
| Grassland | 138 | 0 | 0 | 0 | 3 | 0 | 0 | 135 |
| Agriculture | 718 | 0 | 0 | 0 | 1 | 0 | 0 | 717 |
| Wetland | 12 | 0 | 0 | 0 | 0 | 0 | 0 | 12 |

(Row labels under "LULC before Action")

**Figure 3:** RL-based land use conversion showing the LULC changes in 1000 steps of running the trained PPO model in the real-world environment. The algorithm changes most of the pixels to wetlands as it results in the lowest runoff. The pixels containing urban and wetland areas were left unaltered because of the prior assumption that the LULC couldn't be changed.

## 3. Benefits of an RL-based Sociohydrology Framework

An RL-based sociohydrology framework can offer a wide array of benefits. For example, such a framework can be *transferable*. The core components of the frameworks themselves do not need to change, even though their types can change depending on the problem at hand. For example, in the case study presented here, the Actor creates different LULC scenarios associated with some reward. On a different problem, the Actor might be involved in a different activity, or the reward mechanism might be different; however, the core structure involving the Actor's action and the reward mechanism can stay intact. Thus, the framework can be transferred to a different problem altogether. On the other hand, adding more representativeness to the Actor's functions and having more robust approaches to rewarding can make the framework *generalizable*.

*Multiscale* process simulation has gained popularity in ML-based hydrology. A multiscale model used as an Actor and an appropriate reward mechanism can make the framework work across multiple scales. The framework can also be *scalable*, given the vast popularity of RL across different disciplines and the recent advances related to computational power and efficient pipelines for scalability. One of the major challenges in water resources management is the lack of observed data. Typically, traditional scenario design involves using long-term records. However, in the case of RL, since scenarios themselves coevolve with the other system components, RL will be associated with more *robustness* in a data-scarce setup and will use data more efficiently. This is a significant benefit of RL, as it learns on the go by interacting with the environment.

The issue of *nonstationarity* has been widely acknowledged in recent literature (Milly et al., 2008). We argue that RL is designed to represent nonstationarity and even *tipping points* seamlessly. Learning on the go will lead to learning nonstationary changes. Furthermore, unlike in traditional scenario analysis tasks, RL will have the added advantage of representing nonstationarity across different system components in a coupled manner, which gives it a unique edge over traditional approaches. Along with nonstationarity, for the same reason (i.e., learning on the go), RL will also be able to capture *emergent phenomena*, which is impossible to do through preconceived scenarios, as done traditionally. With drastic changes and tipping points in the environmental systems, proper consideration and representation of emergent phenomena is paramount. Last but not least, an RL-based sociohydrology framework will provide a platform well-suited for analyzing the impacts of a wide range of management interventions, which can add significant value to water management.

## 4. Future Research Directions

Since we are almost at the inception stage of RL use in water management, there are several avenues where coordinated efforts need to be made to make RL more mainstream and leverage the strengths of the framework. For example, we need *foundation models* for RL-based sociohydrology problems. This will ensure we design comprehensive frameworks for solving a wide range of sociohydrology problems by efficiently integrating a variety of information from a wide range of sources, which is much needed, given the complexities inherent in water resources management and the challenges involved in studying human-water interactions. This should be a community effort, collectively accumulating the strengths and perspectives of researchers and practitioners from many disciplines and ensuring responsible use.

Proper *rewarding mechanisms* are essential for meaningful RL implementation, and more work is needed in this area. These reward functions can either be scale-dependent or scale-invariant, depending on the problem at hand. The very nature of the rewarding mechanisms can also vary from problem to problem. A nice trade-off would be to design a set of rewards for each type of problem. For example, understanding LULC impacts on runoff can have a set of reward functions catering to that specific type of problem. Libraries created for this wide range of reward functions will be very useful.

Future research can be directed toward making RL frameworks for sociohydrology more *realistic* and *physically consistent*, which can be achieved through several different means. For example, the policy network can be updated in the *online mode*, meaning that we do not need to use a fixed pre-trained/pre-calibrated model representing the Actor. Instead, the model itself can be updated with the Critic. The framework can also include *causal representation*, as used in causal reinforcement learning (e.g., Gasse et al., 2021). Related to causality, *physics-based constraints* can be added to the model structure or in the loss function to incorporate more realism (e.g., Pokharel et al., 2023). *Expert knowledge* can also be incorporated into these frameworks to add more realism.

Proper characterization and reduction of *uncertainty* is of paramount importance in water resources management, and adding more realism to the system will arguably reduce the epistemic uncertainty. An RL framework can also be used to address *deep uncertainty*, but unlike in scenario-based *robust decision-making*, where competing scenarios are stress-tested, several competing scenarios can co-evolve subject to different constraints imposed. This will lead to a proper characterization of uncertainty in the decisions.

A tool developed based on RL needs to be useful, easily adaptable, and customizable to the *users*. Often, new developments do not easily trickle down to practice, and bridging this gap is important. The tools need to be developed with user preference in mind. The recent advances in generative AI and large language models (LLMs) can be leveraged to create interactive platforms acting as wrappers on the RL framework for better interpretability and ease of use. Using expert knowledge in tool design is one of many ways in which practitioners and policymakers can be involved in the *co-development* process. These tools can be developed with proper consideration of *algorithmic fairness*, something that is crucial in the context of water resources management to avoid unfair judgments. Last but not least, these frameworks can also transcend the *boundaries of sociohydrology* (even though sociohydrology has a fairly wide scope) and incorporate a wide array of other disciplines or sub-disciplines.

## Conflict of Interest

The authors express no conflicts of interest.